\documentclass[journal,transmag]{IEEEtran}
\usepackage{cite}

\usepackage{graphicx}

\ifCLASSINFOpdf
\else
\fi
\newcommand{\ie}{\textit{i}.\textit{e}.}
\newcommand{\eg}{\textit{e}.\textit{g}.}

\begin{document}
%
\title{Learning-from-Observation System\\Considering Hardware-Level Reusability} 


\author{\IEEEauthorblockN{Jun~Takamatsu\IEEEauthorrefmark{1},~\IEEEmembership{member,~IEEE},
Kazuhiro~Sasabuchi\IEEEauthorrefmark{1},~\IEEEmembership{member,~IEEE},
Naoki~Wake\IEEEauthorrefmark{1},~\IEEEmembership{member,~IEEE}, \\
Atsushi~Kanehira\IEEEauthorrefmark{1},~\IEEEmembership{member,~IEEE}, and
Katsushi~Ikeuchi\IEEEauthorrefmark{1},~\IEEEmembership{Life Fellow,~IEEE}}
\IEEEauthorblockA{\IEEEauthorrefmark{1}Applied Robotics,
Microsoft, Redmond, WA 98052 USA}
\thanks{Manuscript received ? ?, ????; revised ? ?, ????. 
Corresponding author: J. Takamatsu (email: jun.takamatsu@microsoft.com).}}

\markboth{Journal of \LaTeX\ Class Files,~Vol.~14, No.~8, August~2015}%
{Shell \MakeLowercase{\textit{et al.}}: Bare Demo of IEEEtran.cls for IEEE Transactions on Magnetics Journals}
%



\IEEEtitleabstractindextext{%
\begin{abstract}

Robot developers develop various types of robots for satisfying users' various demands. Users' demands are related to their backgrounds and robots suitable for users may vary.  If a certain developer would offer a robot that is different from the usual to a user, the robot-specific software has to be changed. On the other hand, robot-software developers would like to reuse their developed software as much as possible to reduce their efforts. 
We propose the system design considering hardware-level reusability. For this purpose, we begin with the learning-from-observation framework. This framework represents a target task in robot-agnostic representation, and thus the represented task description can be shared with various robots. When executing the task, it is necessary to convert the robot-agnostic description into commands of a target robot. To increase the reusability, first, we implement the skill library, robot motion primitives, only considering a robot hand and we regarded that a robot was just a carrier to move the hand on the target trajectory. The skill library is reusable if we would like to the same robot hand. Second, we employ the generic IK solver to quickly swap a robot. We verify the hardware-level reusability by applying 
two task descriptions to two different robots, Nextage and Fetch.
\end{abstract}

}

\maketitle

\IEEEdisplaynontitleabstractindextext

%
\IEEEpeerreviewmaketitle

\section{Introduction}
%
%
%
%

\IEEEPARstart{R}{obot} developers develop various types of robots, such as the bipedal robot and mobile manipulators for satisfying users' various demands. If we consider manipulation aspects, there are various types: a single-arm robot or a dual-arm robot and the degrees of freedom (DOF) of an arm is from 5 DOF (\eg, HSR, Toyota) to 7 DOF (\eg, Fetch Mobile Manipulator, Fetch Robotics). Some robots have additional DOF on their waist.

Users' demands are related to their backgrounds and robots suitable for users may vary. 
If a certain developer would offer a robot that is different from the usual to a user, the robot-specific software has to be changed. On the other hand, robot-software developers would like to reuse their developed software as much as possible to reduce their efforts. It is desirable to satisfy those two conflicting demands.

We have developed a robot system based on a {\em Learning-from-observation} framework~\cite{ikeuchi2021semantic}. In the system, we only need to demonstrate a target task to reproduce the same task by a robot. Thus, even users without robotics knowledge can have the robot perform tasks that they desire. Unlike similar frameworks, such as learning-from-demonstration and imitation learning~\cite{schaal1999imitation,schaal2003imitation,billard2008robot,asfour2008imitation,dillmann2010advances,akgun2012keylfd}, our system first recognizes a task as a symbolic task sequence (such as grasp, pick up, place, and release) and then extracts parameters that are pre-defined in each task, from the demonstration. Hereafter, we regard a task as one action of grasp, etc., and a whole process as a task sequence. These task representations are robot-agnostic. If we would prepare the converter that converts such robot-agnostic representations into the target robot commands, 
our system adopts a target robot with a small effort.

In this paper, we would like to propose a system design considering reusability. We describe it as hardware-level reusability. First, we implement each task only considering a robot hand and we regard a robot as just a carrier to move the hand on the target trajectory. 
We refer to the implemented tasks as skills and the set of skills as a skill library. The difficulty of the implementation depends on the task. To avoid the difficulty, we employ reinforcement learning (RL). Even though the hand hardware is changed, the reward function, which is derived from the characteristics of the task, is reusable. Given the target trajectory by the task, deciding a robot's body motion is so-called inverse kinematics (IK). We keep the reusability of various robots by using the general IK solver, the body-role division~\cite{SasabuchiRAL2021}, 

In this paper, we assume that the environment between in demonstration and in robot execution is not dramatically changed. The demonstration includes a hint for collision avoidance and a robot can execute the target task by following the demonstration. Of course, we admit a slight difference in the environment in robot execution; the system can absorb the difference using sensor feedback.   

\section{Related Work}

Efforts to increase reusability for robots, such as {\em Robot Operating System} (ROS)~\cite{Quigley2009ROSAO} and {\em OpenRTM}~\cite{openrtm}, have been conducted thus far. These two middlewares follow a so-called {\em subsumption architecture}~\cite{Brooks}. In this architecture, the robot program consists of several modules. The modules are properly connected and communicate with each other to execute a robot. These two middlewares achieved the reusability by 1) unifying the format of communication and 2) providing means of communication (\eg, publisher and subscriber in ROS). Switching between low-level modules (\eg, modules to output sensor reading) is very easy. In high-level modules, it is often necessary to take into account the individual characteristics of the robot hardware. 
For example, to move a mobile robot on the floor, we can define de-facto standard robot command (\eg, a pair of velocity and angular velocity) and combinations of modules (the occupancy grid map~\cite{MoravecICRA1985} and Monte Carlo localization~\cite{Dellaert-1999-14893}), and so on. These two middlewares provide open-source modules for various robots.

In manipulation aspects, the control strategy of the manipulator differs from situations, such as simple position control, impedance control~\cite{Hogan1984}, and machine-learning-based control~\cite{JIN201823}. But to achieve the manipulation at a minimum, the end-effector must be brought to the desired position without colliding with any obstacles. To bring to the desired position, it is necessary to decide on the robot joint to satisfy the target end-effector position, \ie, to solve IK. Many proposals and implementations are available, such as~\cite{Beeson2015}. For collision avoidance, the random-sampling-based motion planning, such as the probabilistic roadmap method~\cite{KavrakiITRA1996}, Rapidly-exploring Random Trees~\cite{LavalleAFR2000} and their variants, are often used as a de-facto standard. And ROS also provides the unified motion planning framework, {\em MoveIt}. These softwares require the target positions of the end-effector as input to achieve task sequences. That requires another software to generate these positions. Our main concern is to develop the software to generate the positions considering its reusability.

\section{System Design for Hardware Reusability}

\subsection{Definition of Tasks}

We assume that the task sequence starts with grasping a target object, continues by manipulating it, and ends to release it. Thus, we can define the tasks as a set of grasp, manipulation, and release. Furthermore, manipulation is split into tasks based on the constraint of the target object motion, \eg, feasible displacement~\cite{ikeuchi2021semantic}. For example, in the case of {\em picking up} an object, the feasible translation of the object in the beginning is limited to the upward of the support surface, and that in the end is all space. By using the Kuhn-Tucker theory~\cite{kuhn1957linear}, the translation can be classified into 10 types. In the same way, rotational displacement can be classified into 13 types. By considering the possible transitions of the types, the tasks can be defined from the corresponding transitions. For the details, please see~\cite{ikeuchi2021semantic}. Note that there are two categories of constraints: physical constraint and semantic constraint (when wiping, a cloth can be physically moved above but its vertical motion be semantically constrained for wiping).

\subsection{Tasks and Skill Library}

In each task, an object's motion is calculated to satisfy the constraint. The displacement in the constraint-free space defines the movement and the system obtains such displacement from a demonstration. On the other hand, the constraint itself comes from the environment. A vision system estimates the constraint parameter on the robot execution. These two parameters are defined as {\em skill parameters}. 

We implemented the skill library in advance. Each skill corresponds to each task and given the skill parameters and robot states (\eg, joint state and tactile), each skill outputs the target hand configuration in each control step. Some tasks, such as PTG3 (opening/closing a drawer) and PTG5 (opening/closing a revolute door), were implemented as the reinforcement-learning (RL) agent; when training, the RL agent learns the motion not to violate the constraint (\eg~avoiding infeasible displacement) using force feedback. Grasp was also implemented as the RL agent~\cite{SaitoHumanoids2022}. Some tasks were implemented by a manual program and PTG13 (placing an object) was implemented by a manual program with force feedback. In short, given the skill parameters and robot states, skills return the target hand configuration in each control step.

\subsection{Hand Motion to Body Motion under Hardware-Level Reusability}

Generally, IK converts hand motion to body motion. Given the target hand configuration, the general IK solver, such as~\cite{Beeson2015}, solves the body configuration that satisfies the hand configuration, by minimizing the difference between the target configuration and the configuration obtained by the forward kinematics under some joint configuration. Generally, there are many solutions that satisfy the target hand configuration and IK sometimes outputs unexpected posture. 
The unexpected posture is likely to appear as the DOF of the robot becomes larger. 

To avoid the unexpected posture, we follow the idea of the body-role division~\cite{SasabuchiRAL2021} and employ Labanotation~\cite{GuestBOOK1970,IkeuchiIJCV2018}. In Labanotation, the pose of each limb is represented by 26 discretized directions and there is a margin to achieve a certain target Labanotation pose. The Labanotation pose is obtained from the demonstration (one of the skill parameters).
To apply the Labanotation constraint, we manually set the joint angles to each Labanotation pose. We solve IK as long as the joint angles do not go outside the specified range. 
Note that for arms with joints that can rotate infinitely, the Labanotation pose can be set to limit the angle of the joint at the elbow to the positive direction to reduce unnecessary movement.

The number of degrees of Labanotation constraints should depend on the DOF of the robot. For example, in the case of Nextage with the fixed waist joint, DOF is six in a single-arm manipulation. Even if we do not apply the Labanotation constraint, the unexpected solution of the IK seldom occurs. On the other hand in the case of Fetch with a lifter, DOF is eight. Thus, we apply the full Labanotation constraint, \ie, the constraint of four joints (two from an upper arm direction and two from a lower arm direction), especially when generating initial posture\footnote{We can apply Labanotation constraint in every IK solution. Due to the difference between the human body configuration and the robot's configuration, a Labanotation pose obtained from the demonstration may not be suitable every time. Thus, we apply the constraint when deciding the initial posture and use the current posture as the initial guess in IK after that.}. 


\section{Implementation}

\subsection{Testbed}

\begin{figure}
    \centering
    \includegraphics[width=\linewidth]{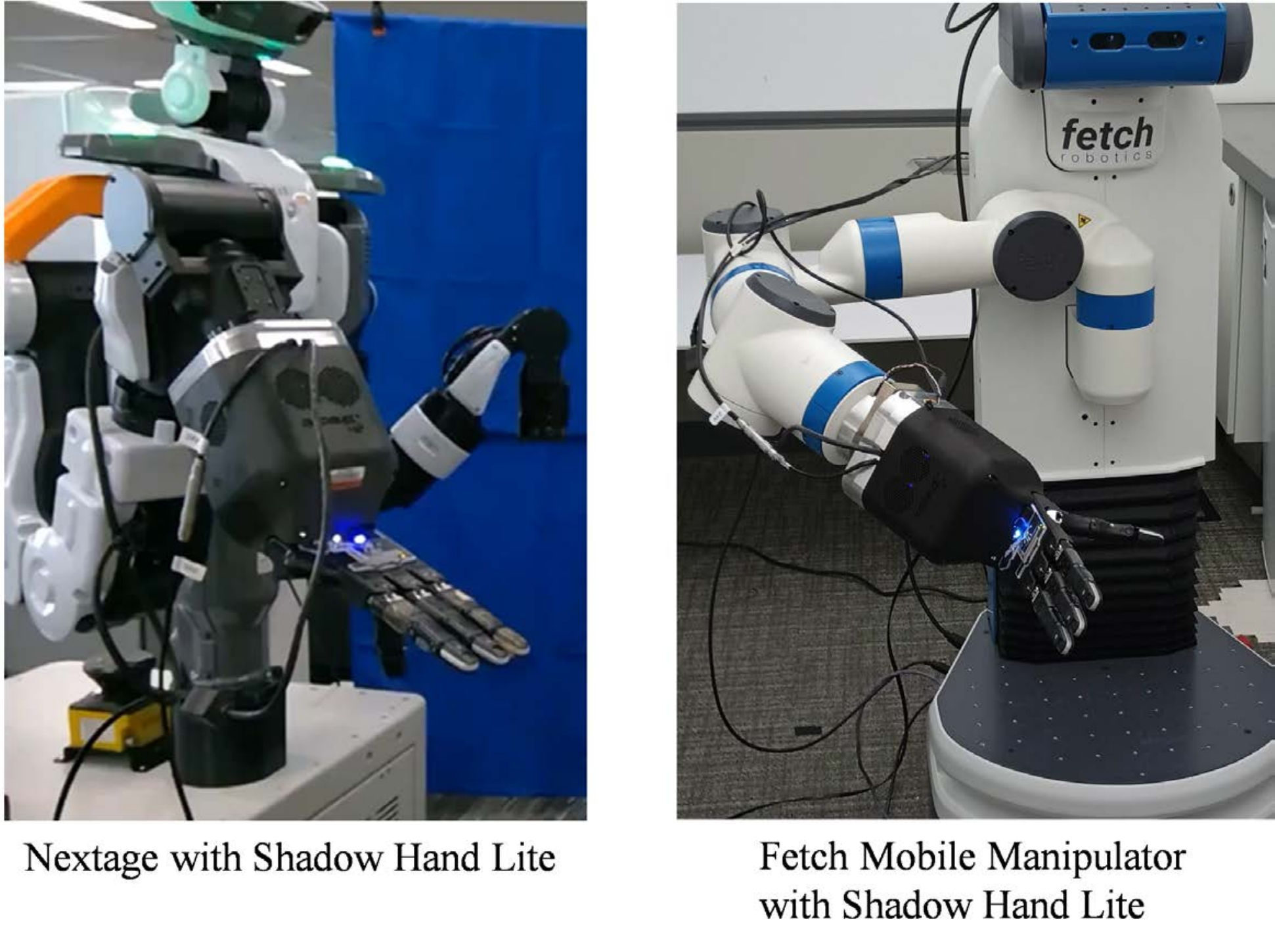}
    \caption{Two testbed robots: Nextage, Kawada Robotics and Fetch Mobile Manipulator, Fetch Robotics. They are equipped with Shadow Hand Lite, Shadow Robotics, as an end effector.}
    \label{fig:testbed}
\end{figure}

We use two testbed robots shown in Figure~\ref{fig:testbed}. One is Nextage, Kawada Robotics. It has two arms and each arm has six DOF. It also has one DOF in the waist (rotation around the vertical axis). In this paper, we use only the right arm, not the left arm or waist, to perform the task. The right arm is equipped with Shadow Hand Lite, Shadow Robotics, as an end effector. Nextage is equipped with a stereo camera to observe an environment in a 3-dimensional space. 

The other is Fetch Mobile Manipulator, Fetch Robotics. It has one arm with 7 DOF, 1 DOF in the waist (moving up and down), and 2 DOF in a mobile base. It is also equipped with Shadow Hand Lite. We do not use a mobile base. It is equipped with an RGB-D camera, Primesense Carmine 1.09, to observe an environment. 

\subsection{Observation of Demonstration}

We use an RGB-D camera, Azure Kinect, Microsoft, to observe the demonstration. We put an AR marker to align the orientation of the robot coordinates and the demonstration coordinates. By doing this, the robot executes the same task sequence by achieving the demonstrated displacement with collision avoidance. Note that we adjust the displacement for placing an object since the displacement is different from that of the demonstration due to the difference in grasping position due to execution variation. 

\begin{figure}
    \centering
    \includegraphics[width=\linewidth]{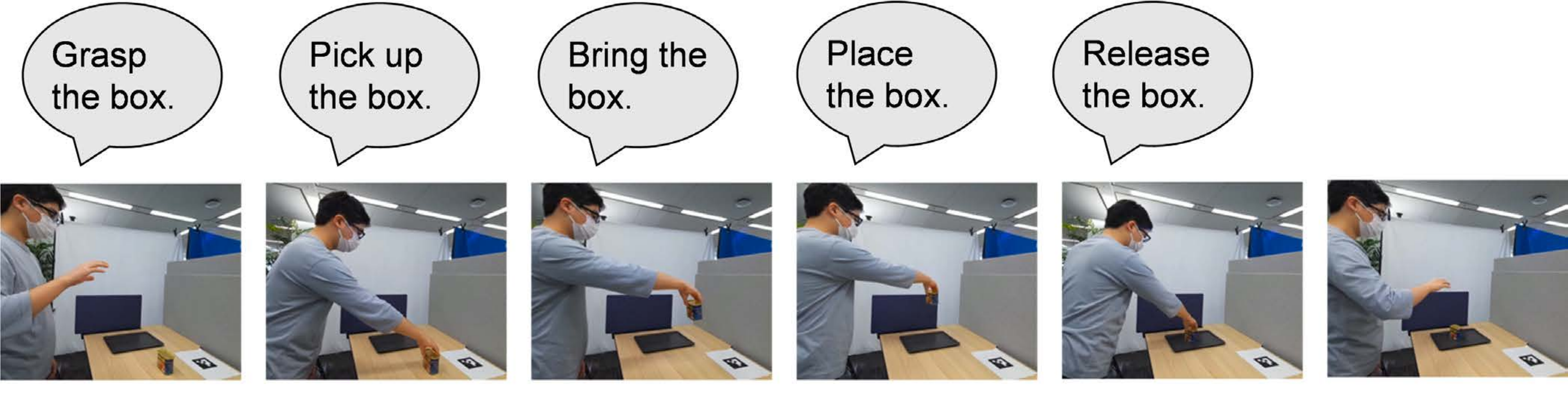}
    \caption{Overview of the demonstration}
    \label{fig:demonstration}
\end{figure}

In the demonstration, we use a stop-and-go manner. By stopping the motion, the demonstrator manually segments each task in the task sequence. When carrying an object, we add via points to avoid collisions. And in each segment, we utter the explanation of the task (\eg,~``grasp the box" and ``pick up the box"). 

From the explanation and the demonstration image sequences, the task and its parameters can be estimated~\cite{wake2021learning}. 
First, the type of each task is estimated. Then the skill parameters of each task are estimated by observing the task sequence again. Generally, the hand position and Labanotation pose at the beginning and the end of the task is included in the skill parameters. 
Additionally, in grasping, the grasping type and approach direction are also the skill parameters. In PTG13, the surface normal (assuming an upright direction in the robot coordinates, \ie, z-axis) is also the skill parameter.

\subsection{Skill Library}

In this paper, we implemented the following skills: Grasp with three grasp types (passive-force grasp, active-force grasp, and lazy grasp), Release, PTG11 (pick up), STG12 (bring), and PTG13 (place). We used the implementation of the three grasping skills in~\cite{SaitoHumanoids2022}. For PTG11 and STG12, the end position of the hand was first obtained by adding the displacement obtained in the demonstration to the starting position. Via points are added according to the length of the trajectory. After solving the IK in each point, robot movement is obtained by linear interpolation in a joint space. The orientation of the object, corresponding to the orientation of the hand, is kept in PTG11, while the upright direction of the object is kept in STG12.

As described above, it is necessary to adjust the displacement in PTG13. First, we decide the position for starting landing from the demonstration parameters (\eg, the middle position in the demonstration). Then, a robot moves its hand at a small step size and checks the hitting condition using force feedback. If the hitting is detected, a robot terminates PTG13 and goes to the next task. For the force feedback, we used the effort values of the finger joints of the hand. When hitting, the values are changed. 

\subsection{Skill Parameters on Execution}

In the current system, the parameters on the execution are related to grasp only. The necessary parameters were defined in~\cite{SaitoHumanoids2022} and were the positions of the virtual fingers and the orientation of the hand when grasping. We defined these values in target objects whose shape is represented in superquadrics~\cite{superquadrics}. We render depth images of objects by changing the shape parameters and view directions. And we train CNN to estimate the necessary parameters from a depth image. For the details, please see~\cite{SaitoHumanoids2022}.

The estimation of the parameters is performed by the following steps. First, we segment an object on a depth image by applying the plane estimation with random sample consensus (RANSAC)~\cite{ransac} to a depth image; we assume that the object is on the planer surface. Then we input the segmented depth image to the trained CNN that outputs the parameters. 

\section{Experiment}

\begin{figure}[t]
    \centering
    \includegraphics[width=\linewidth]{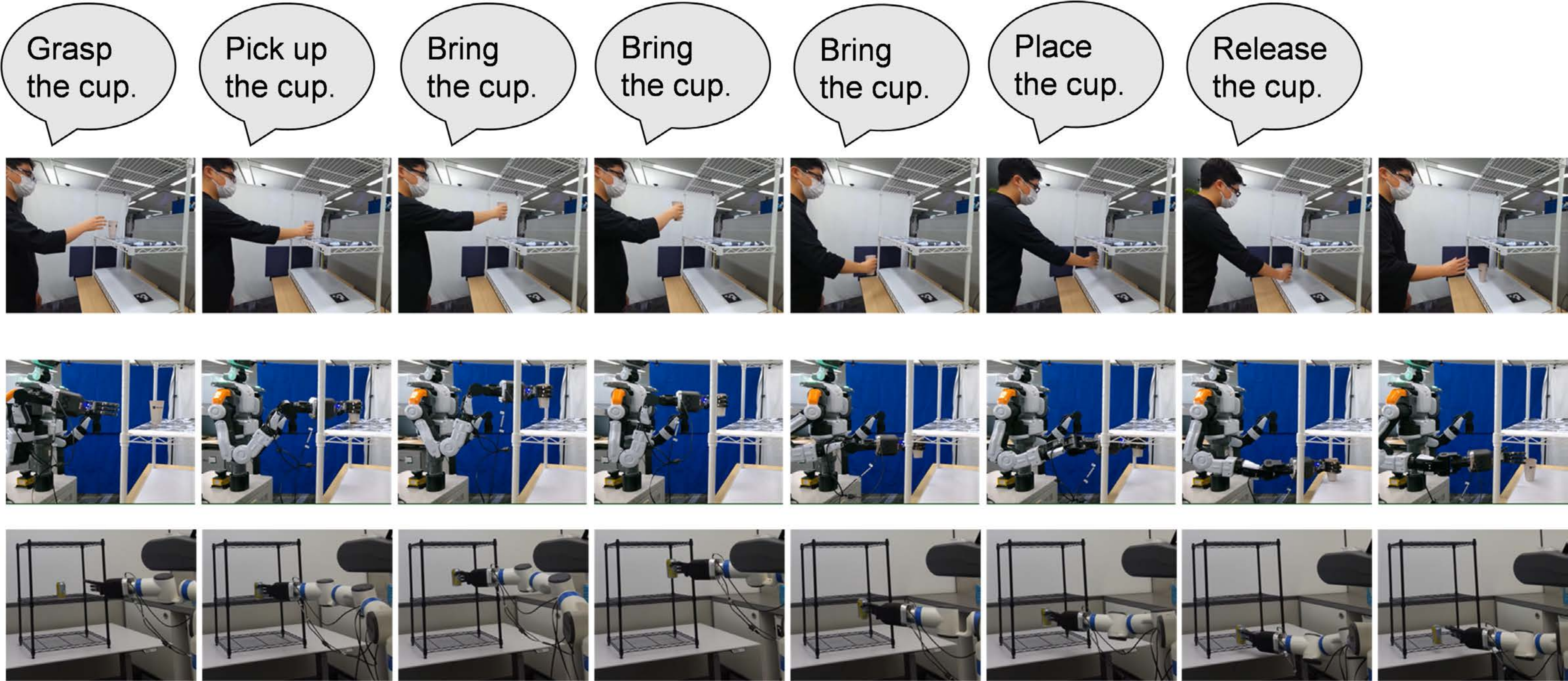}
    \caption{Shelf demo. Top: Human demonstration. Middle: Execution by Nextage. Bottom: Execution by Fetch.}
    \label{fig:shelf_demo}
\end{figure}

To verify hardware-level reusability, we applied the same task sequence to the two robots. We used two demonstrations. 
The first one (referred to as {\em place-on-plate} demo) consists of grasping the box, picking up the box from a table, bringing the box, placing the box on a plate, and releasing the box (See Figure~\ref{fig:demonstration}). The second one (referred to as {\em shelf} demo) consists of grasping the cup, picking up the cup, three repetitions of bringing the cup, and releasing the cup (See Figure~\ref{fig:shelf_demo}). 

\subsection{Place-on-Plate Demo}

\begin{figure}
    \centering
    \includegraphics[width=\linewidth]{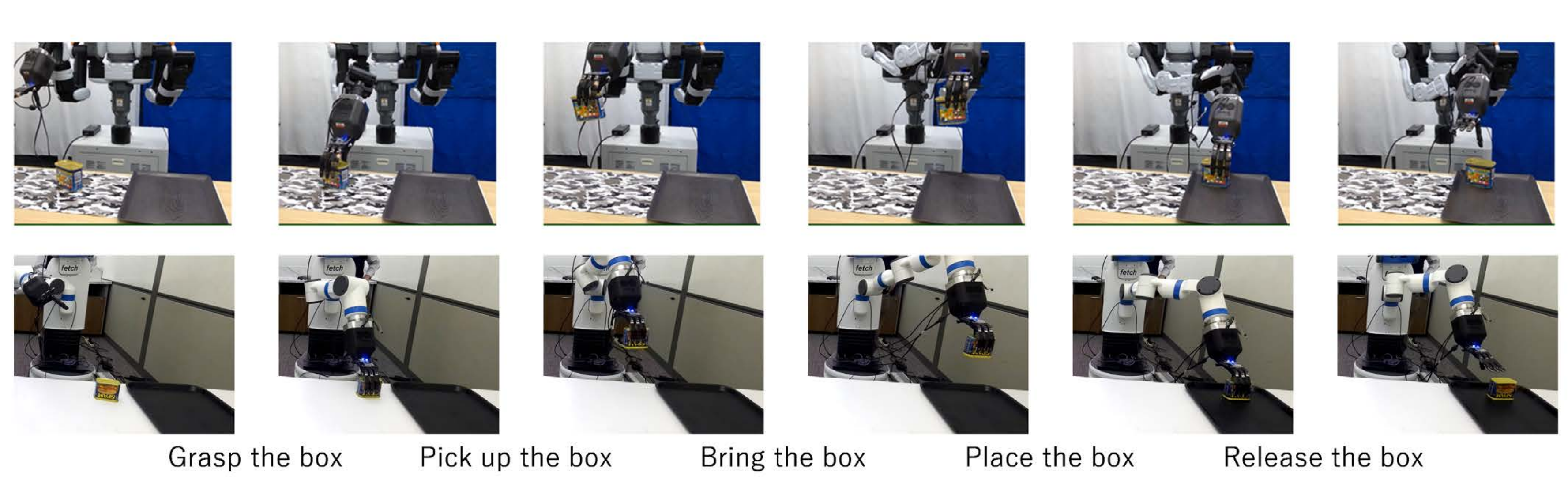}
    \caption{Place-on-plate demo. Top: Execution by Nextage. Bottom: Execution by Fetch. These are the reproduction of the demonstration in Figure~\ref{fig:demonstration}.}
    \label{fig:place_on_plate_demo}
\end{figure}

For performing the place-on-plate demonstration, we first demonstrated the task sequence in front of Azure Kinect. As the result, the task sequence was recognized as grasp, PTG11, STG12, PTG13, and Release. After obtaining the task sequence, the skill parameters were estimated by observing the task sequence again. 
Figure~\ref{fig:place_on_plate_demo} shows the execution by the two robots. By 
executing the task sequence as the same as the observed one, the two robots executed the same task sequence.  

\subsection{Shelf Demo}

Figure~\ref{fig:shelf_demo} shows the results from the demonstration to the execution of the shelf demo by the two robots. The task sequence was recognized as grasp, PTG11, three pieces of STG12, PTG13, and Release. And the skill parameters were estimated by observing the task sequence as similar to the place-on-plate demo. Using the obtained skill parameters and executing the task sequence, the two robots executed the same task sequence. In this demonstration, the demonstrator used three pieces of STG12 to teach the robots the trajectory for avoiding the collision with the shelf and the robots successfully avoided the collisions.

\section{Discussion}

In this paper, we used the same robot hand, Shadow Hand Lite. As the result, we can share the skill library for two robots. It is necessary to implement IK and connection to sensors and a robot only. 
If we would use another robot's hands, the skills that should be changed are Grasp and PTG13. Though we design the state and action in RL for the implementation of Grasp, we believe that the reward function and the training environment can be shared. For PTG13, it is necessary to implement the detection for hitting and that depends on the sensor that a robot is equipped with. But hitting is usually detected by the value change of the sensor. From this discussion, we conclude that the proposed system realizes hardware-level reusability as much as possible.

\section{Conclusion}

We proposed the system design considering hardware-level reusability. For this purpose, we began with the learning-from-observation framework~\cite{ikeuchi2021semantic}. This framework represents a target task in robot-agnostic representation, and thus the task sequences with skill parameters that are obtained from the demonstration can be shared with various robots. When executing the task sequence, it is necessary to convert robot-agnostic representation to commands of a target robot. To increase the hardware-level reusability, first, we implemented the skill library only considering a robot hand and we regarded that a robot was just a carrier to move the hand on the target trajectory. Second, we employed the generic IK solver, body role division~\cite{SasabuchiRAL2021} to quickly swap a robot. We verified the hardware-level reusability by applying two 
demonstrations to two different robots, Nextage and Fetch. The two robots were able to execute the same task sequences.


%

\ifCLASSOPTIONcaptionsoff
  \newpage
\fi



\bibliographystyle{IEEEtran}
\bibliography{paper}
%

\end{document}